\documentclass{llncs}
\usepackage{makeidx}  

\usepackage{graphicx}
\usepackage{algorithm}
\usepackage{algorithmic}

\usepackage{amssymb}
\usepackage{amsmath}
\usepackage{amsfonts}

\begin{document}
\frontmatter          
\pagestyle{headings}  
\mainmatter              
\title{A Learning Framework for Morphological Operators using Counter--Harmonic Mean}
\titlerunning{A Learning Framework for Morphological Operators using Counter--Harmonic Mean}  
%
\author{Jonathan Masci\inst{1}\thanks{This work has been supported by ArcelorMittal, Maizi\`{e}res Research, Measurement and Control Dept., France.} 
\and Jes\'{u}s Angulo\inst{2} \and J\"urgen Schmidhuber\inst{1}}
\authorrunning{Masci et al.} 
%
\tocauthor{Jonathan Masci, Jes\'{u}s Angulo, J\"urgen Schmidhuber}
\institute{IDSIA -- USI -- SUPSI, Manno--Lugano, Switzerland \\
\email{jonathan@idsia.ch},\\ WWW home page:
\texttt{http://idsia/\homedir masci/} \and CMM-Centre de Morphologie
Math\'{e}matique, \\ Math\'{e}matiques et Syst\`{e}mes, MINES
ParisTech, France
}

\maketitle              

\begin{abstract}
We present a novel framework for learning morphological operators using counter-harmonic mean.
It combines concepts from morphology and convolutional neural networks.
A thorough experimental validation analyzes basic morphological operators dilation and erosion, opening and closing, 
as well as the much more complex top-hat transform, for which we report a real-world application from the steel industry.
Using online learning and stochastic gradient descent,
our system learns both the structuring element and the composition of operators.
It scales well to large datasets and online settings.

\end{abstract}

\section{Introduction}
Mathematical morphology (MM) is a nonlinear image processing
methodology based on computing max/min filters in local
neighborhoods defined by structuring
elements~\cite{Serra82,Soille99}. By concatenation of two basic
operators, i.e., the dilation $\delta_{B}(f)$ and the erosion
$\varepsilon_{B}(f)$, on the image $f$, we obtain the closing
$\varphi_{B}(f) = \varepsilon_{B}\left(\delta_{B}(f) \right)$ and
the opening $\varphi_{B}(f) = \gamma_{B}\left(\delta_{B}(f)
\right)$, which are filters with scale-space properties and
selective feature extraction skills according to the underlaying
structuring element $B$. Other more sophisticated filters are obtained by combinations of openings and closings, to address problems such as non-Gaussian denoising, image regularization, etc.

Finding the proper pipeline of morphological operators and
structuring elements in real applications is a cumbersome and time
consuming task. In the machine learning community there has always
been lot of interest in learning such operators, but due to the
non-differentiable nature of the max/min filtering only few approaches have been found to succeed, notably one based on LMS (gradient steepest descent algorithm) for rank filters formulated with the sign function \cite{Salembier92,Salembier93}. 
This idea was later revisited \cite{Pessoa98} in a neural network framework combining morphological/rank filters and linear FIR filters. Other attempts from the evolutionary community (e.g., genetic algorithms~\cite{Harvey96} or simulated annealing~\cite{Wilson93}) use black-box optimizers to circumvent the
differentiability issue. However, most of the proposed approaches do
not cover all operators. More importantly, they cannot learn both the structuring element and the operator, e.g., ~\cite{Nakashizuka10}. 
This is obviously a quite
important limitation as it makes very hard or even impossible the
composition of complex filtering pipelines. Furthermore, such
systems are usually limited to a very specific application and 
hardly generalize to complex scenarios.

Inspired by recent work \cite{Angulo10} on counter-harmonic
mean asymptotic morphology, we propose a novel framework to
learn morphological pipelines of operators. We combine convolutional neural networks (CNN) with a new type of layer that permits complex pipelines through multiple layers 
and therefore extends this models to a Morphological Convolutional Neural Network (MPCNN). 
It extends previous work on deep-learning while making directly applicable all optimization tricks and findings of this field.

Here we focus on methodological foundations and show
how the model learns several operator pipelines,  from
 dilation/erosion to top-hat transform (i.e.,
residue of opening/closing). We report an important real-world
application from the steel industry, and present a sample
application to denoising where operator learning outperforms hand-crafted structuring elements.

%
Our main contributions are:
\begin{itemize}
\item a novel framework where learning of morphological operators and filtering pipelines can be performed using gradient-based techniques, exploiting recent insights of deep-learning approaches;
\item the introduction of a novel PConv layer for CNN, to let CNN benefit from highly nonlinear, morphology-based filters;
\item the stacking of many PConv layers, to learn complex pipelines of operators such as opening, closing and top-hats.
\end{itemize}

\section{Background}
Here we illustrate the foundations of our approach,
introducing the Counter-Harmonic Mean formulation for asymptotic
morphology and CNN. 

\subsection{Asymptotic morphology using Counter-Harmonic Mean}

We start from the notion of counter-harmonic
mean~\cite{Bullen87}, initially used ~\cite{vanVliet04}
for constructing robust morphological-like operators. More recently,
its morphological asymptotic behavior was characterized~\cite{Angulo10}. Let $f(x)$ be a 2D real-valued image, i.e., $f :
\Omega \subset \mathbb{Z}^2 \rightarrow \mathbb{R}$, where $x\in
\Omega$ denotes the coordinates of the pixel in the image domain
$\Omega$. Given a (positive) weighting kernel $w : W \rightarrow
\mathbb{R}_{+}$, $W$ being the support window of the filter, the
\emph{counter-harmonic mean (CHM) filter of order $P$}, $-\infty
\leq P \leq \infty$ is defined by,
\begin{equation}
\kappa_{w}^{P}(f)(x) = \frac{(f^{P+1}\ast w)(x)}{(f^{P}\ast w)(x)} =
\frac{\int_{y \in W(x)} f^{P+1}(y)w(x-y)dy}{\int_{y \in W(x)}
f^{P}(y)w(x-y)dy},
\end{equation}
where $f^{P}$ is the image, where each pixel value of $f$ is
raised to power $P$, $/$ indicates pixel-wise division, and $W(y)$ is
the support window of the filter $w$ centered on point $y$. We note
that the CHM filter can be interpreted as \emph{$P-$deformed
convolution}, i.e., $\kappa_{w}^{P}(f)(x) \equiv (f\ast_{P} w)(x)$.
For $P \gg 0$ ($P \ll 0$) the pixels with largest (smallest) values
in the local neighborhood $W$ will dominate the result of the
weighted sum (convolution), therefore morphological dilation and
erosion are the limit cases of the CHM filter, i.e., $\lim_{P
\rightarrow + \infty} (f\ast_{P} w)(x) =$ $\sup_{y \in W(x)} f(y) =$
$\delta_{W}(f)(x)$, and $\lim_{P \rightarrow - \infty} (f\ast_{P}
w)(x) =$ $\inf_{y \in W(x)} f(y) =$ $\varepsilon_{W}(f)(x)$, where
$W$ plays the role of the structuring element. As  proven earlier~\cite{Angulo10}, 
apart from the limit cases (e.g., a typical order of
magnitude of $5 \leq |P | < 10 $), we have the following behavior:
\begin{eqnarray}
(f\ast_{P} w)(x)\mid_{P\gg 0} & \approx & \sup_{y\in W(x)}\left\{
f(\mathbf{y}) + \frac{1}{P} \log\left( w(x-y)\right) \right\} , \\
(f\ast_{P} w)(x)\mid_{P\ll 0} & \approx &  \inf_{y\in W(x)}\left\{
f(\mathbf{y}) - \frac{1}{P} \log\left( w(x-y)\right) \right\},
\end{eqnarray}
which can be interpreted, respectively, as the nonflat dilation
(supremal convolution) and nonflat erosion (infimal convolution)
using the structuring function $b(x) = \frac{1}{P}\log\left(
w(x)\right)$. By using constant weight kernels, i.e., $w(x) = 1$ if
$x\in W$ and $w(x) = 0$ if $x\notin W$, and $|P| \gg 0$, we just
recover the corresponding flat structuring element $W$, associated
to the structuring function $\mathfrak{w}(x) = 0$ if $x\in W$ and
$\mathfrak{w}(x) = -\infty$ if $x\notin W$.

From a precise morphological viewpoint, we notice that for
finite $P$ one cannot guarantee that $(f\ast_{P} w)(x)$
yields exactly a pair of dilation/erosion, in the sense of
commutation with max/min~\cite{Serra82,Soille99}. Consequently,
\emph{stricto sensu}, we can only name them as pseudo-dilation
($P\gg 0$) and pseudo-erosion ($P\ll 0$). The asymptotic cases of
the CHM filter can be also combined to approximate opening and
closing operators, i.e.,
\begin{equation}\label{OpeningClosing}
    \left\{
    \begin{array}{lll}
      \left( (f\ast_{-P} w) \ast_{P} w \right)(x) & \xrightarrow{P\gg 0} &  \gamma_{W}(f)(x),\\
      \left( (f\ast_{P} w) \ast_{-P} w \right)(x) & \xrightarrow{P\ll 0} &
      \varphi_{W}(f)(x).
    \end{array}
    \right.
\end{equation}


\subsection{Convolutional Neural Networks}
CNN are hierarchical models
alternating two basic operations, convolution and subsampling,
reminiscent of simple and complex cells in the primary visual cortex
\cite{hubel:1968}. 
Their main characteristic
 is that
they exploit the 2D structure of images via weight sharing,
learning a set of convolutional filters. 
Certain CNN
scale well to real-sized images
and excel in many object recognition
\cite{ciresan:2011a,ciresan:2011b,ciresan:2011c,masci:2012ijcnn} and
segmentation \cite{Ciresan:2012f,Turaga:2010} benchmarks. We refer
to a state-of-the-art CNN as depicted in Figure~\ref{fig:cnn}. It
consists of several basic building blocks briefly explained here:
\begin{figure}[!btp]
\begin{center}
\includegraphics[width=0.7\linewidth]{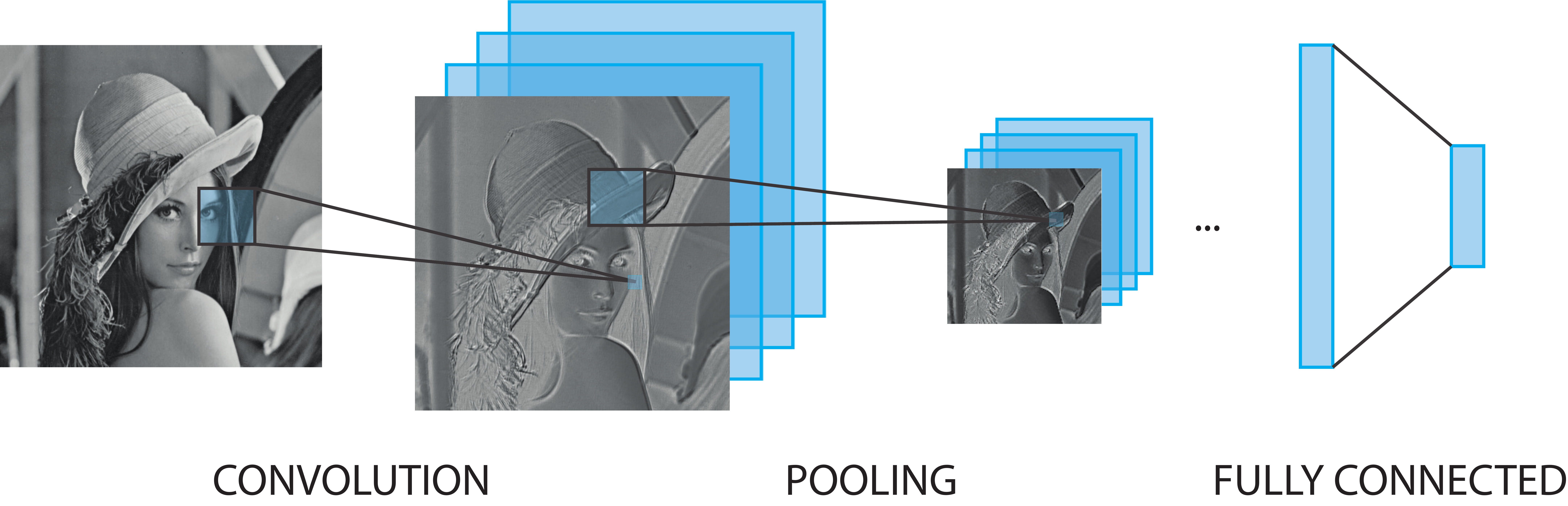}
\caption{A schematic representation of an CNN.
Convolutional and pooling layers are stacked below the fully connected
layers used for classification.}
\label{fig:cnn}
\end{center}
\end{figure}
\begin{itemize}
\item \textit{Convolutional Layer}: performs a 2D filtering between input images
$\left\{f_i \right\}_{i\in I}$ and a bank of filters $\left\{w_k
\right\}_{k\in K}$, producing another set of images $\left\{h_j
\right\}_{j\in J}$ denoted as maps. Input-output correspondences are
specified through a connection table $CT$ (inputImage $i$, filterId $k$,
outputImage $j$). Filter responses from
inputs connected to the same output image are linearly combined.
This layer performs the following mapping:
$
h_{j}(x) = \sum_{i,k \in CT_{i,k,j}} (f_{i} \ast w_{k})(x),
$
where $\ast$ indicates the 2D valid convolution. 
Then, a
nonlinear activation function (e.g., $tanh$, $logistic$, etc.) is
applied to $h_j$.
Recently $relu$ activations have been found to excel. They are the units
we  use for all our models. A $relu$ unit operates as $relu(x) = \max(lb, \min(x, ub))$.
It is common choice to use $0$ as lower bound ($lb$) and $\infty$ as upper bound ($ub$).
\item \textit{Pooling Layer}: down-samples the input images by a constant factor keeping a value (e.g. maximum or average) for every non overlapping subregion of size $p$ in the images.
Max-Pooling is generally favorable, as it introduces small invariance
to translation and distortion, leads to faster convergence and
better generalization \cite{scherer:2010}.
\item \textit{Fully Connected Layer}: this is the standard layer of a multi-layer network.
It performs a linear multiplication of the input vector by a weight
matrix. 
\end{itemize}

Note the striking similarity between the max-pooling layer and a dilation transform. 
The former is in fact a special case of dilation,
with a square structuring element of size $p$ followed by downsampling (sampling one out of every $p$ pixels). 
Our novel layer, however, does not any longer
limit the pooling operation to simple squares, but allows for  a much richer repertoire of structuring elements fine-tuned
for given tasks. This is what makes MCNN so powerful.

\section{Method}
Now we are ready to introduce the novel morphological layer based on
CHM filter formulation, referred to as {\bf PConv} layer.
For a single channel image $f(x)$ and a single filter $w(x)$ the
PConv layer performs the following operation
\begin{equation}
PConv(f; w, P)(x) = \frac{(f^{P+1} \ast w)(x)}{(f^P \ast w)(x)} = (f
\ast_{P} w)(x)
\end{equation}
It is parametrized by $P$, a scalar which controls the type of
operation ($P < 0$ pseudo-erosion, $P>0$ pseudo-dilation and $P=0$
standard linear convolution), and by the weighting kernel $w(x)$,
where the corresponding asymptotic structuring function is given by
$\mathfrak{w}(x) = \log\left( w(x)\right)$. Since this formulation is
differentiable we can use gradient descent on these parameters.

The gradient of such a layer is computed by back-propagation~\cite{Werbos:74,LeCun:85}.
In minimizing a given objective function $L(\theta; X)$, where $\theta$
represents the set of parameters in the model, back-propagation applies
the chain rule of derivatives to propagate the gradients down to the input
layer, multiplying them by the Jacobian matrices of the traversed layers.
Let us introduce first two partial results of back-propagation
\begin{equation}
\Delta_U (x) = \frac{f(x)}{(f^P \ast w)(x)}; \quad \Delta_D(x) =
\frac{-f(x) \cdot (f^{P+1} \ast w)(x)}{(f^P \ast w)(x)}.
\end{equation}
The gradient of a PConv layer is computed as follows
\begin{eqnarray}
\frac{\partial L}{\partial w} & = & \tilde{f}^{P+1} * \Delta_U + \tilde{f}^P \ast \Delta_D  \\
\frac{\partial L}{\partial P} & = & f^{P+1} \cdot \log(f) \cdot
(\frac{f}{f^P \ast w} \ast \tilde{w}) + f^P \cdot \log(f) \cdot
(\Delta_D \ast \tilde{w})
\end{eqnarray}
where $\tilde{f}, \tilde{w}$ indicate flipping along the two dimensions and $\cdot$ indicates element-wise multiplication.
The partial derivative of the PConv layer with respect to its input (to back-propagate the gradient) is instead
\begin{equation}
\frac{\partial (f \ast_{P} w)(x)}{\partial f} = \Delta_U(x) +
\Delta_D(x).
\end{equation}

Learning the top-hat operator requires a
{\em short-circuit} in the network to allow for subtracting the input
image (or the result of an intermediate layer) from the output of a
filtering chain. For this purpose we introduce the {\bf
AbsDiffLayer} which takes two layers as input and emits the absolute
difference between them. Partial derivatives can still
be back-propagated.

\subsection{Learning Algorithm}
Minimizing a PConv layer is a non-convex,
highly non-linear operation prone to local convergence. 
Deep-learning findings tell us that stochastic gradient descent is the most effective algorithm to train such complex models. 
In our experiments we use its full online version where
weights are updated sample by sample. The learning rate decays during training.
To further avoid bad local minima we use a momentum term.
For the opening/closing tasks we also alternate between learning $P$,
keeping $w$ fixed, and vice-versa. 
This is common in online dictionary learning and sparse coding. 
We also constrain $w \ge 0$.

\section{Experiments}
We thoroughly evaluate 
our MCNN 
on
several tasks. First we assess the quality of dilation/erosion
operators, which require a single PConv layer.
This gives a good measure of how well training can be performed
using the CHM derivation. 
Then we investigate a two-layer network 
learning openings/closings. This is already a challenging task hardly
covered in previous approaches.

We then learn the top-hat transform for  a very challenging steel industry  application. Using $2$ filters per layer we
learn to simultaneously detect two families of defects without
resorting to multiple training. 
Our implementation allows for learning  multiple filters for
every layer, thus producing a very rich set of filtered maps. Subsequent convolutional layers can learn highly nonlinear
embeddings. (We believe that this will also dramatically improve 
segmentation capabilities of such models to be
investigated in future work.)
We also show that a simple CNN does {\em not} learn well pipelines of morphological operators. This is
actually expected \emph{a priori} due to the nature of
conventional convolutional layers,
and shows the value of  our novel PConv layer.

As final benchmark we consider denoising. Our MCNN shows the superiority of
operator learning over hand-crafting structuring elements for
non-Gaussian (binomial and salt-and-pepper) noise. We also show that
our approach performs well on total variation (TV) approximation
for additive Gaussian noise.

In all our experiments we use stochastic gradient descent and a filter size of
$11 \times 11$ unless otherwise stated. The per-pixel mean-squared error loss
(MSE) is used.

\subsection{Learning dilation and erosion}
\begin{figure}[htbp]
\begin{center}
\includegraphics[width=.8\linewidth]{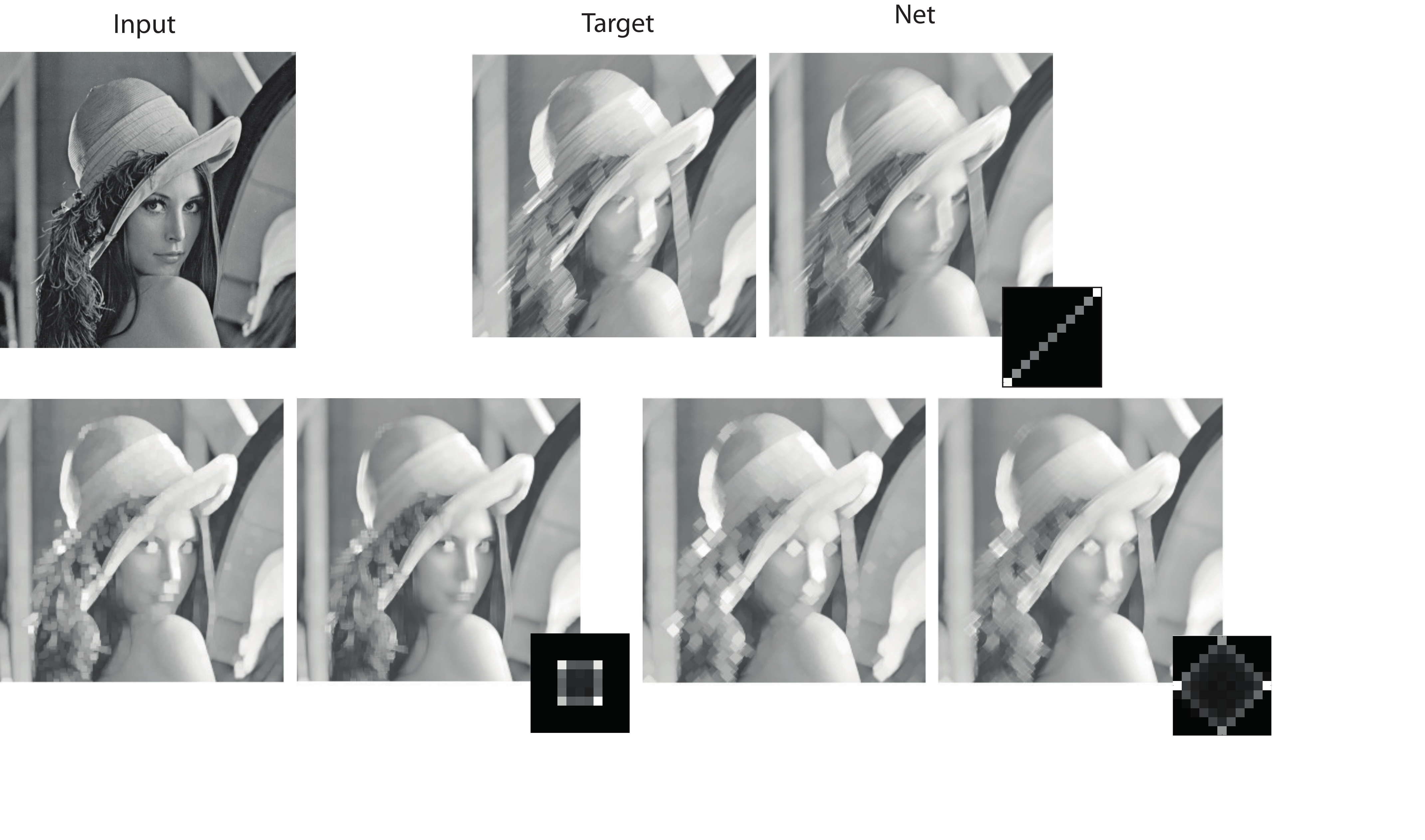}
\caption{Examples of learning a dilation with three different
structuring elements. The target and net output are slightly smaller
than the original image due to valid convolution. The obtained
kernel $w(x)$ for each case is also depicted.} \label{fig:dilate}
\end{center}
\end{figure}

\begin{figure}[htbp]
\begin{center}
\includegraphics[width=.8\linewidth]{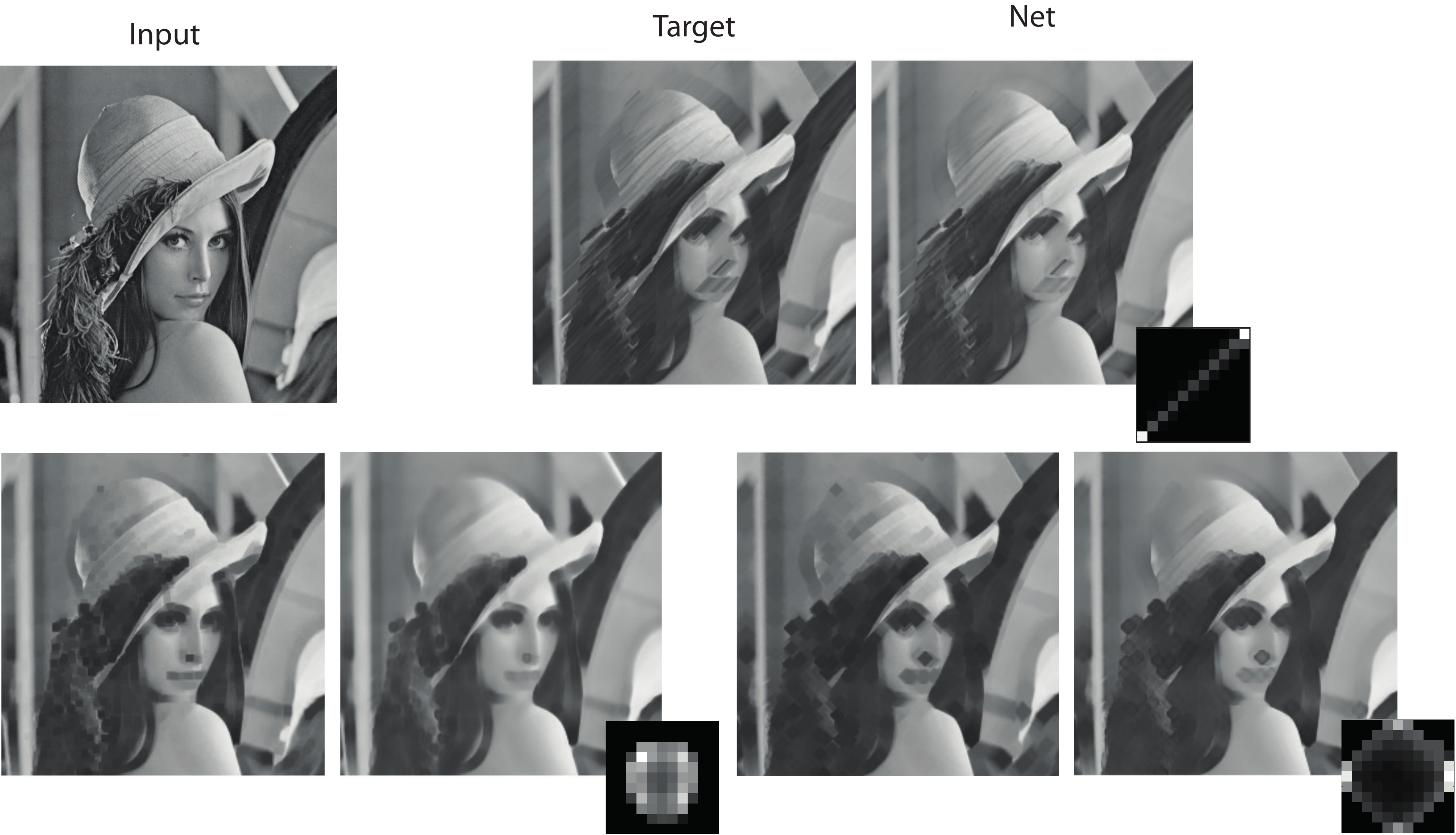}
\caption{Examples of learning an erosion with three different
structuring elements along with the learned kernel $w(x)$.} \label{fig:erode}
\end{center}
\end{figure}

In this first set of experiments we create a dataset as follows: for
every input image $f_i$ we produce a target image $t_j$ using a
predetermined flat structuring element $B_k$ and a predetermined
operator: $t_j = \delta_{B_k}(f_i)$ or $t_j =
\varepsilon_{B_k}(f_i)$. We train until convergence.
Overfitting is not an
issue in such a scenario. The net actually learns the true function underlying the
data. In fact, for an image of $512\times512$ and a structuring
element with support of $11\times 11$ there are $502^2$
patches, way more than the $11^2$ elements in the structuring
element. A CNN with equal topology fails,
producing mainly Gaussian blurred images, illustrating
the need for a PConv layer to handle this kind of nonlinearities.
%
Figure \ref{fig:dilate} shows the results of a
dilation with three structuring elements: a line of $15$ pixels and $45^\circ$, a square of size $5$ pixels and a diamond of
side $5$ pixels. Figure \ref{fig:erode} shows similar results for
the erosion transform. 
%
%
Note that the learned weighted kernels $w(x)$
are not exactly uniformly equal to $1$. The corresponding
morphological structuring function $\mathfrak{w}(x)$, obtained after
applying the logarithm on the weights, produces a rather flat shape.
In practice, we observed that learning an erosion is slightly more
difficult than learning the dual dilation. This is related to the
asymmetric numerical behavior of CHM for $P>0$ and $P<0$.
Nevertheless, in all cases  the learned operator has
excellent performance.

\subsection{Learning opening and closing}
In this set of experiments we train our system to learn openings
$\gamma_{B_{k}}(f_i)$ and closings $\varphi_{B_{k}}(f_i)$. Learning such functions is extremely difficult.
To the best of our knowledge, we are the first to do this in
a flexible and gradient-based framework without any prior.
For instance, in classical approaches~\cite{Salembier92} or
more recent ones~\cite{Nakashizuka10}, the operator needs to be fixed \emph{a priori}.

\begin{figure}[htbp]
\begin{center}
\includegraphics[width=.7\linewidth]{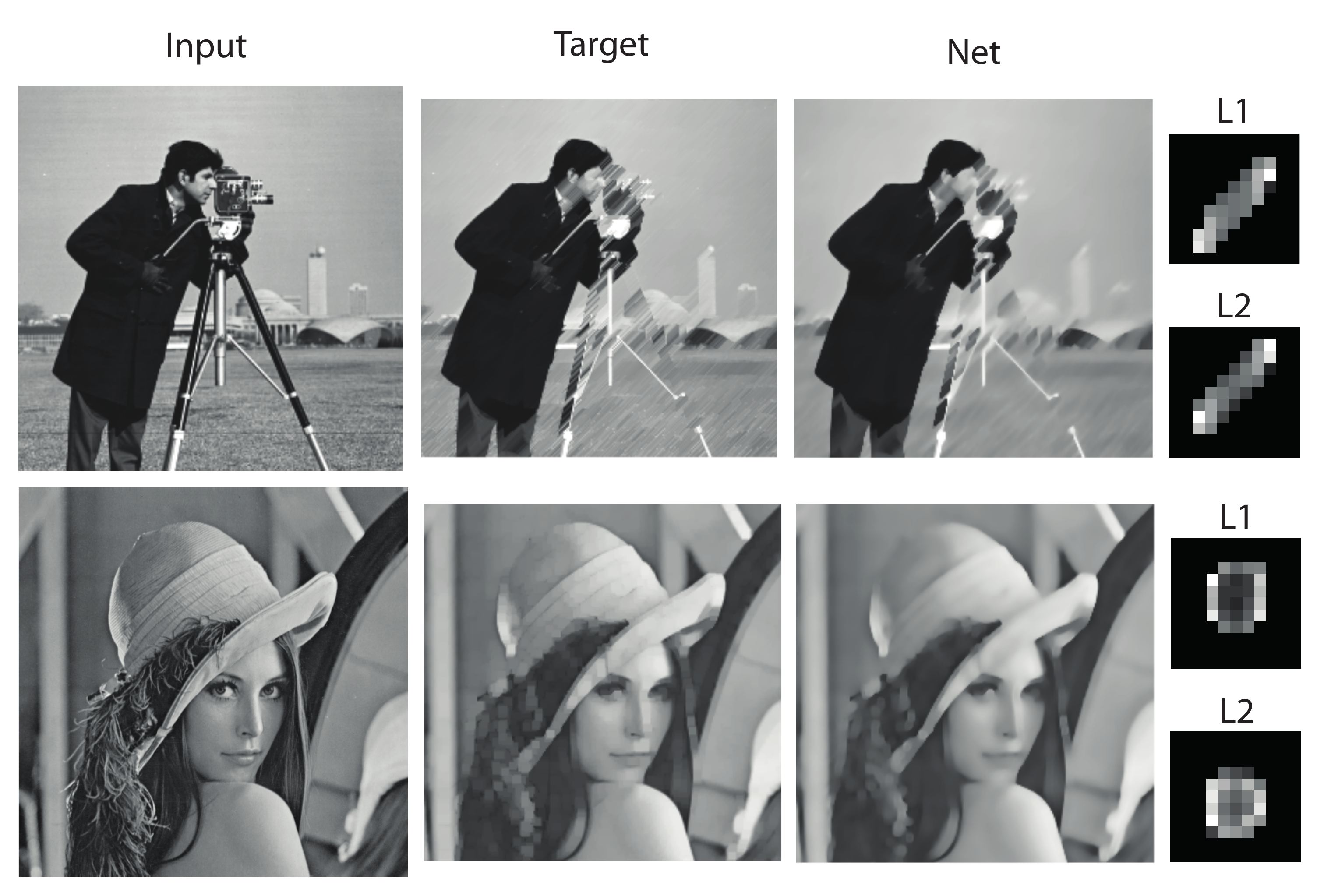}
\caption{{\em Top}: an example of learning a closing operator where a line of length $10$ and orientation $45^{\circ}$ is used. {\em Bottom}: opening with a square structuring element of size $5$. The network closely matches the output and almost perfectly learns the
structuring elements in both PConv layers.} \label{fig:openclose}
\end{center}
\end{figure}

Figure \ref{fig:openclose}-top shows an example of a closing with a line of
length $10$ and an orientation of $45^{\circ}$, whereas Figure
\ref{fig:openclose}-bottom shows an example of an opening with a square of size
$5$. In both cases, the obtained kernel for the first $L1$ and
second $L2$ PConv layers are depicted. We see that the associated
structuring element is learned with a good approximation. On the
other hand, however, we also start to see that learning a flat
opening/closing is remarkably hard and that the network output
starts to be slightly ``blurry''. Why? On
the one hand, the obtained values of $P$, e.g., in the closing
$P_{L1} = 6.80$ and $P_{L2} = -8.85$, in the opening $P_{L1} =
-7.64$ and $P_{L2} = 7.07$ are in the interval of asymptotically unflat
behavior. On the other hand, they are not totally symmetric. We intend to
further study these issues  in ongoing work.
%
%
\subsection{Learning top-hat transform}
Delegating the learning to a neural network allows for easily
constructing complex topologies by linking several simple modules.
We recall that the white top-hat is the residue of the opening, i.e.,
$\varrho^{+}_{B_k}(f_i) = f_i - \gamma_{B_k}(f_i)$, and the black
top-that the residue of the closing, i.e., $\varrho^{-}_{B_k}(f_i) =
\varphi_{B_k}(f_i) - f_i$. Thus, to learn a top-hat
transform we introduce the {\bf AbsDiff} layer. It takes two layers
as input and emits their absolute difference in output.
Backpropagation is performed as usual.

Top-hat is particularly relevant in real applications
such as steel surface quality control. It  is a
powerful tool for defect detection. Here we first show that our
framework can learn such a transform. Increasing the number of filters per
layer from $1$ to $2$, we show that our MCNN
is also much more powerful when jointly learning two
transforms.
\begin{figure}[!h]
\begin{center}
\includegraphics[width=.5\linewidth]{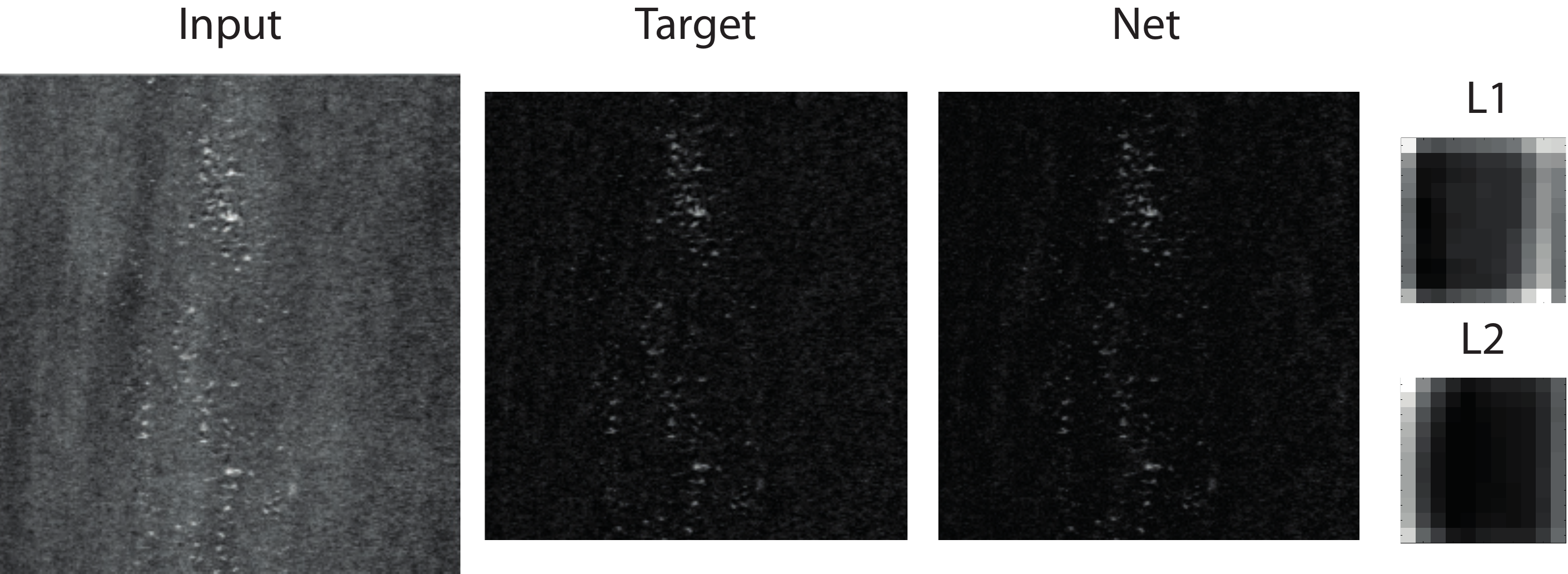}
\caption{Learning a top-hat transform. The defected input image has
bright spots to be detected. The network performs almost
perfectly on this challenging task.} \label{fig:top-hat}
\end{center}
\end{figure}
Figure \ref{fig:top-hat} shows the results for a single top-hat. We create our
training set by applying a white top-hat $\varrho^{+}_{B}$, where
$B$ is a disk of size $5$ pixels. This operator extracts only the
structures of size smaller than $B$ and brighter than the
background. We clearly see that the network performs almost
perfectly. To further assess the advantages
of a PConv layer over a conventional convolutional layer, we also
train a CNN with identical topology.
The discrepancy between the two models in terms
of losses (MSE) is large: we have 1.28E-3 for our MCNN and 1.90E-3 for
the CNN. More parameters are required for a CNN to reach better performance.
This clearly establishes  the added
value of our MCNN.

\begin{figure}[!htb]
\begin{center}
\includegraphics[width=.9\linewidth]{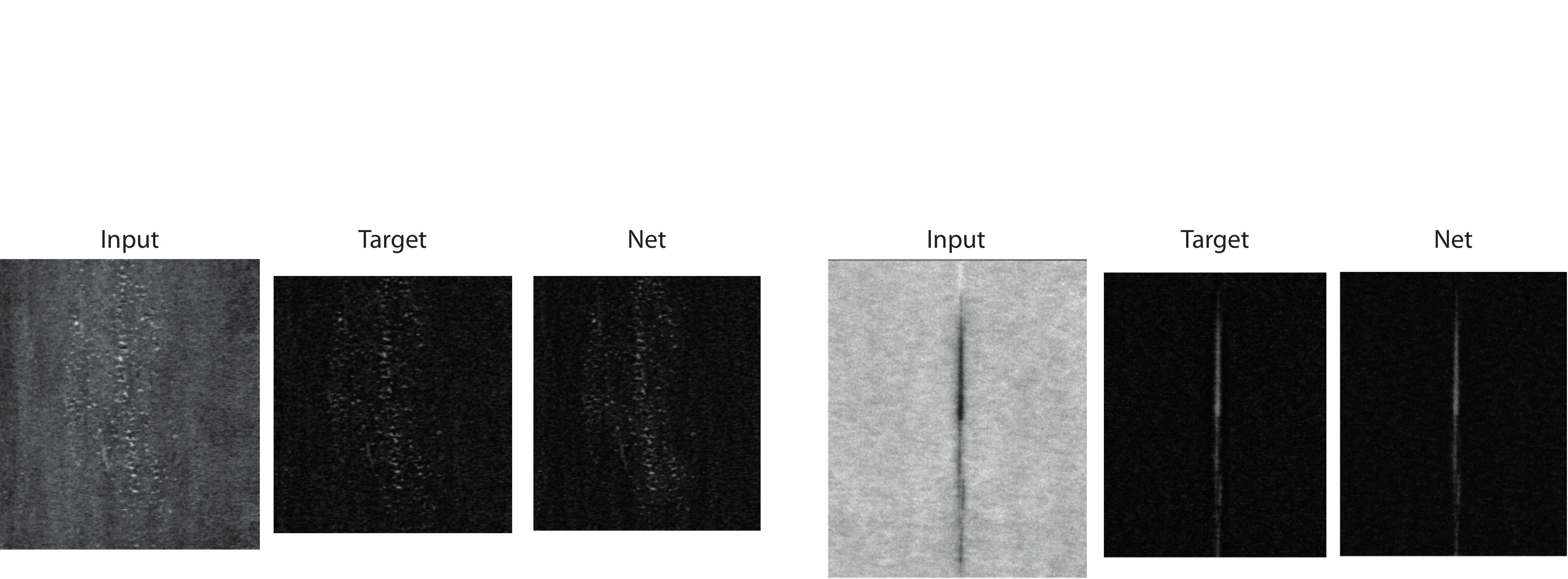}
\caption{Learning two top-hat transforms. On the left, bright spots need to be detected. On the right, a dark vertical line. The network performs almost
perfectly in this challenging task.} \label{fig:top-hat-2strel}
\end{center}
\end{figure}

The steel industry requires many top-hat operations. 
Tuning them one by one is a cumbersome process. Furthermore,
several models' outputs need to be considered to obtain the final
detection result. Figure \ref{fig:top-hat-2strel} shows that by simply
increasing the number of filters per layer we can simultaneously
learn two top-hat transforms and address this problem. 
We learn a white top-hat
with a disk of size $5$ and a black top-hat with a line of size $10$
and orientation of $0^{\circ}$. A convolutional layer is used to combine the output of
the two operators. 
The architecture is as follows: 2 PConv layers, Conv layer, AbsDiff layer with the input. 
The network is almost
perfect from our viewpoint.
This opens the possibility of
using such a setup in more complex scenarios where several
morphological operators should be combined. This is of great
interest in multiple class steel defect detection. 
%

\subsection{Learning denoising and image regularization}
Finally we
compare our MCNN to conventional
morphological pipelines in the denoising task. 
Morphological filters are recommended for non-Gaussian denoising. The
purpose of this evaluation, however, is not to propose a novel noise
removal approach, but  to show the advantages of a learnable
pipeline over a hand-crafted one.
%
\begin{figure}[!h]
\begin{center}
\includegraphics[width=.8\linewidth]{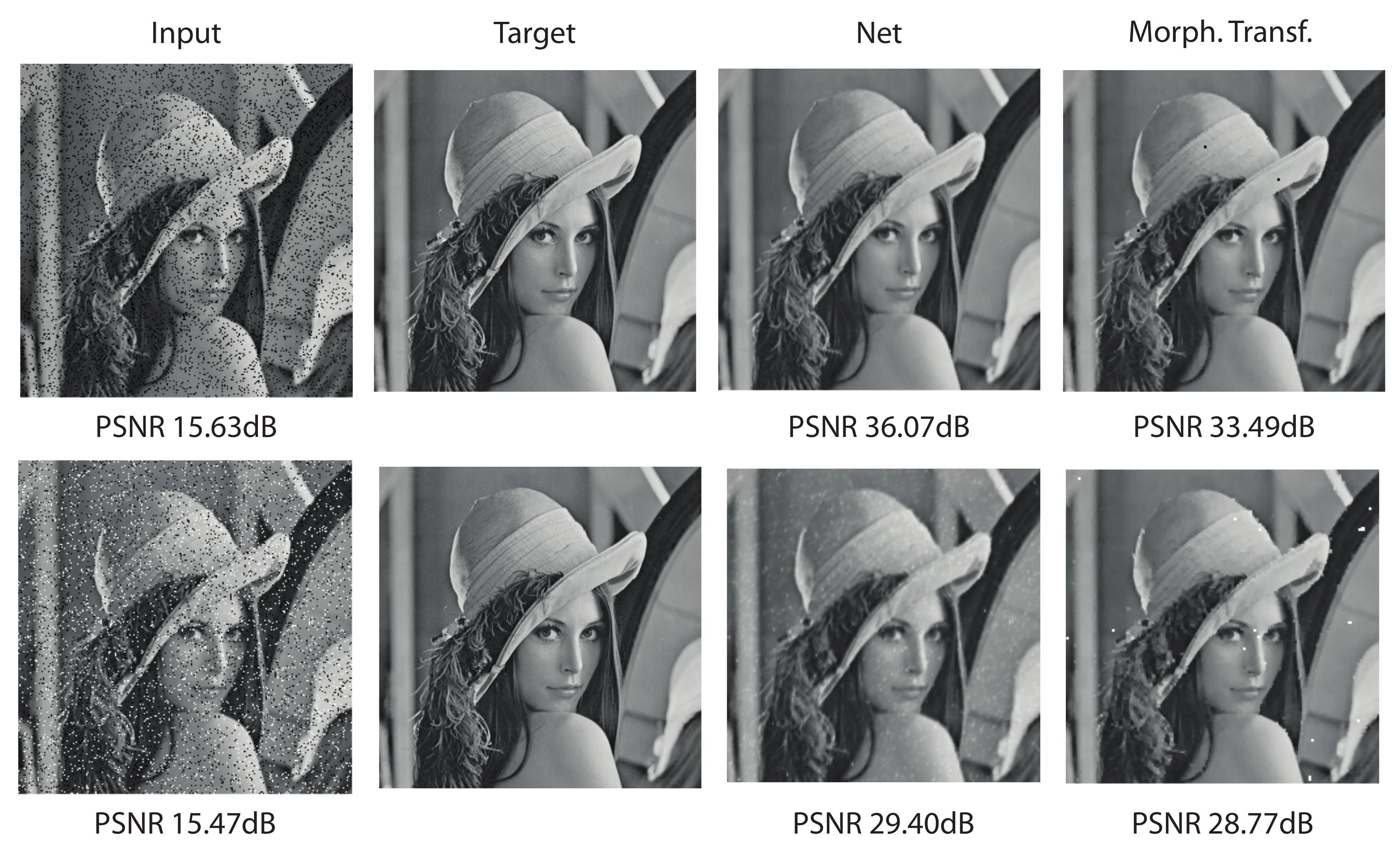}
\caption{{\em Top}: Binomial noise removal task. The learned nonlinear operator
performs better than the hand-crafted one. Learning uses noisy and original images---there is no prior on
the task. {\em Bottom}: Salt'n'pepper noise removal task. Even here, the learned operator performs 
better than the corresponding morphological one.} \label{fig:binsaltpepper_denoise}
\end{center}
\end{figure}
\vspace{-.1cm}
We start with {\em binomial noise} where $10$\% of the image pixels
are switched-off. The topology in this case is:
$2$ PConv layers and filter size of $5\times5$. We compare to a
closing with a square of size $2$, empirically found to deliver
best results. We make the task even harder by using larger-than-optimal support. 
Training is performed in 
fully on-line fashion. While the target images are kept fixed, the
input is generated by adding random noise sample by sample. So the
network never sees the same pattern twice. Figure
\ref{fig:binsaltpepper_denoise}--top compares the two
approaches, and shows the noisy image. We see that learning substantially improves
the PSNR measure.

We continue with an even more challenging task, a $10\%$
salt'n'pepper denoising. The network is made of $4$
PConv layers, a very long pipeline. We compare to an opening
with a square of size $2\times2$ on a closing with the same
structuring element. Training  follows the same
protocol as the one for the binomial noise. Images are generated
online. This creates a possibly infinite dataset with very small
memory footprint. Figure
\ref{fig:binsaltpepper_denoise}--bottom shows results. Although we
can observe some limitations of our approach, it still exhibits the best
PSNR also in this application.

\begin{figure}[htbp]
\begin{center}
\includegraphics[width=.7\linewidth]{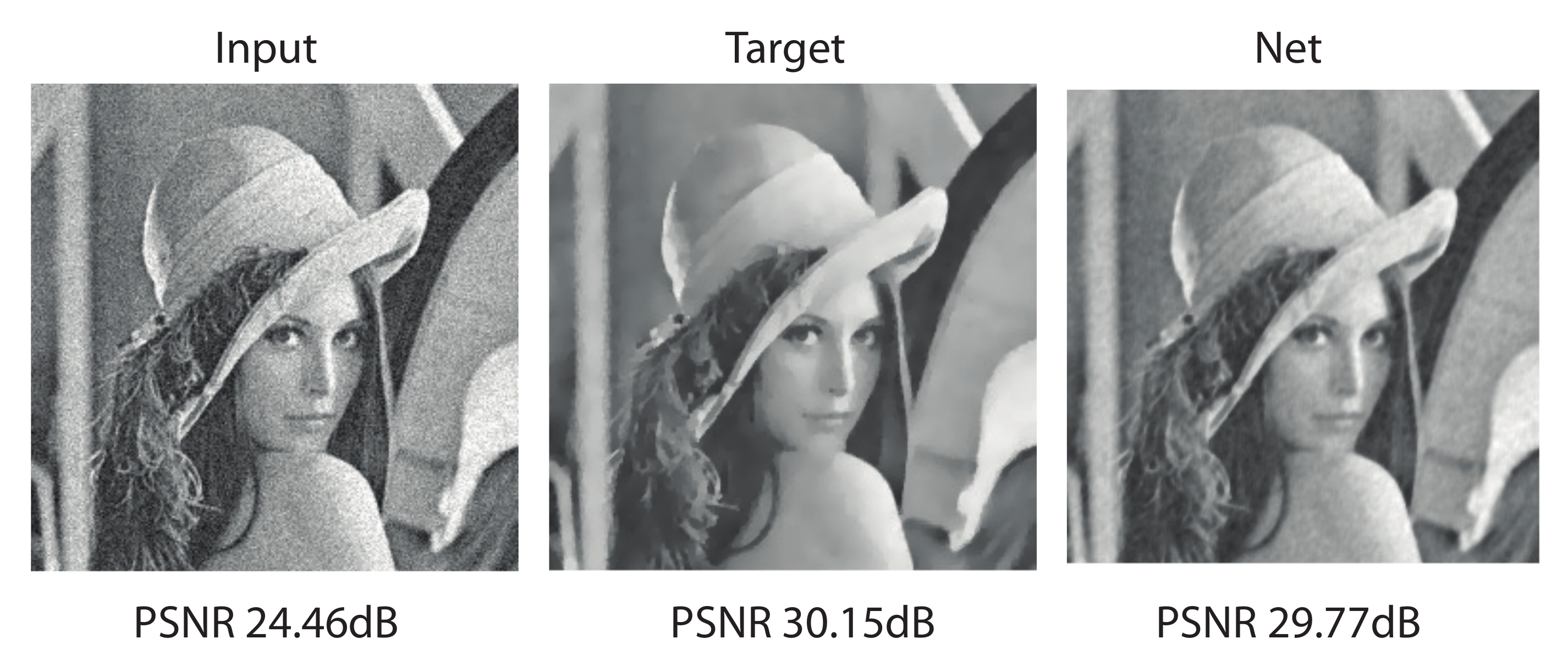}
\caption{Total Variation (TV) task. The network has to learn to approximate the TV output (target) by means of averaging
two filtering pipelines.}
\label{fig:tv_denoise}
\end{center}
\end{figure}

%
Finally we consider the case of {\em total
variation (TV)} restoration from an image corrupted by $6$\%
additive Gaussian noise. Morphological filtering does not excel at 
this task. A MCNN is trained to learn the mapping from noisy image to TV restored image. 
How well can it approximate any target
transformation with a pseudo-morphological pipeline? The architecture is composed of $2$ PConv layers with $2$ filters each
plus an averaging layer. Results are shown in Figure
\ref{fig:tv_denoise}.

\vspace{-.1cm}

\vspace{-1.5cm}

\section{Conclusion and Perspectives}
Our MCNN for learning morphological operators is based on a novel PConv convolutional layer and inherits all the benefits of gradient-based deep-learning algorithms.
It can easily learn complex topologies and operator chains such as white/black top-hats
and we showed its application to steel defect detection.
In future work we intend to let MCNN simultaneously learn
banks of morphological filters and longer filtering pipelines.


%
%
\bibliographystyle{splncs03}
\bibliography{references}

\end{document}